\documentclass[10pt,twocolumn,letterpaper]{article}

\usepackage{iccv}              
\usepackage[accsupp]{axessibility}
\usepackage{multirow}
\usepackage{amsfonts}
\usepackage{algorithm}
\usepackage{algpseudocode}
\usepackage{bm}
\PassOptionsToPackage{table}{xcolor}
\usepackage{listings}
\usepackage[misc]{ifsym}
\usepackage{tcolorbox}
\tcbuselibrary{most}
\tcbset{
  aibox/.style={
    width=\textwidth,
    top=10pt,
    colback=white,
    colframe=black,
    colbacktitle=black,
    enhanced,
    center,
    attach boxed title to top left={yshift=-0.1in,xshift=0.15in},
    boxed title style={boxrule=0pt,colframe=white,},
  }
}
\newtcolorbox{AIbox}[2][]{aibox,title=#2,#1}
\definecolor{iccvblue}{rgb}{0.21,0.49,0.74}
\usepackage[pagebackref,breaklinks,colorlinks,allcolors=iccvblue]{hyperref}

\def\paperID{7827} 
\def\confName{ICCV}
\def\confYear{2025}

\title{Large Multi-modal Models Can Interpret Features in Large Multi-modal Models}

\author{Kaichen Zhang$^{1,2}$, Yifei Shen$^{2,3}$, Bo Li$^{1,2}$, Ziwei Liu$^{1,2}$\textsuperscript{\Letter}\\
$^{1}$S-Lab, NTU, Singapore, $^{2}$LMMs-Lab Team\\
{\tt\small \{zhan0564,libo0013,ziwei.liu\}@e.ntu.edu.sg} \\
$^{3}$Microsoft Research Asia\\
{\tt\small yshenaw@connect.ust.hk}
}

\begin{document}
\twocolumn[{%
   \renewcommand\twocolumn[1][]{#1}%
   \maketitle
   \vspace{-30pt}
   \begin{center}
    \centering
    \includegraphics[width=0.95\textwidth]{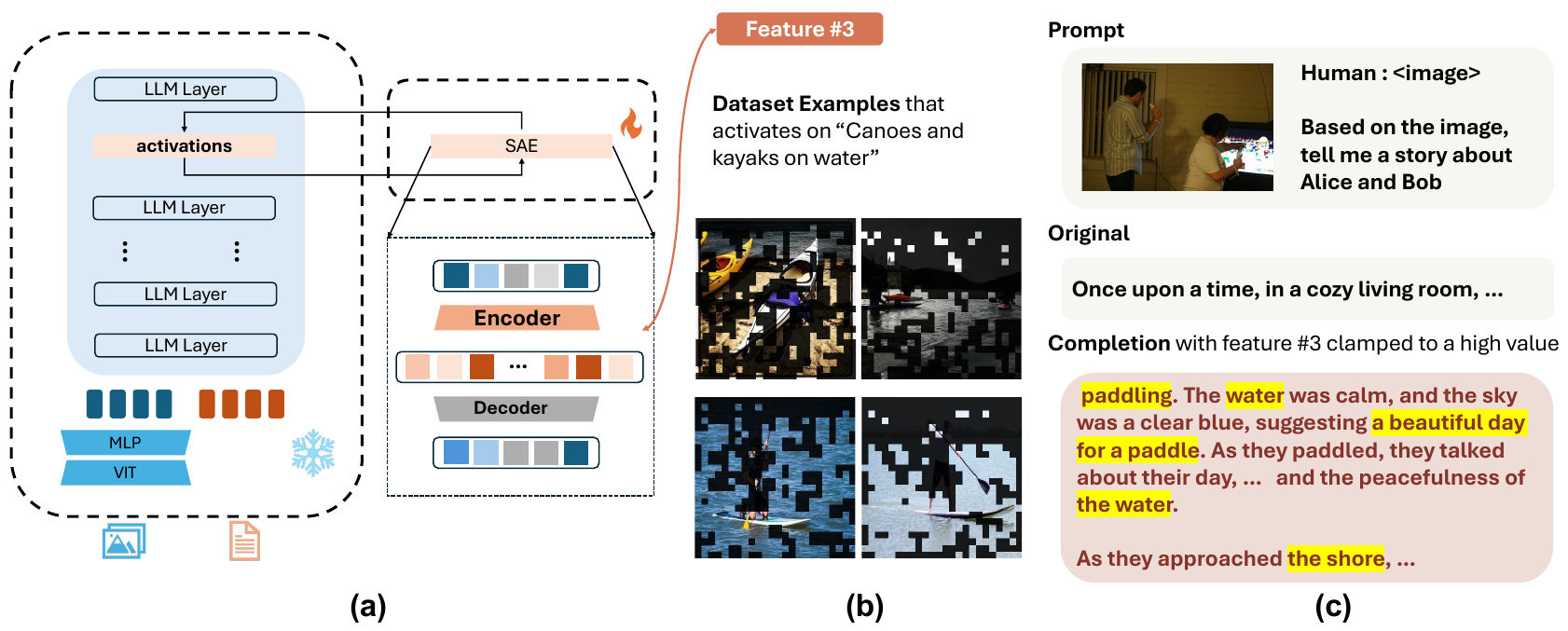}
    \vspace{-10pt}
    \captionof{figure}{\textbf{a)} The Sparse Autoencoder (SAE) is trained on LLaVA-NeXT data by integrating it into a specific layer of the model, with all other components frozen. \textbf{b)} The features learned by the SAE are subsequently interpreted through the proposed auto-explanation pipeline, which analyzes the visual features based on their activation regions. \textbf{c)} It is demonstrated that these features can be employed to steer the model's behavior by clamping them to high values.}
    \label{fig:sae-banner}
   \end{center}%
  }]

\renewcommand{\thefootnote}{\fnsymbol{footnote}}
\footnotetext[1]{\textsuperscript{\Letter}Corresponding author.}
\renewcommand*{\thefootnote}{\arabic{footnote}}
  
\begin{abstract}
    Recent advances in Large Multimodal Models (LMMs) lead to significant breakthroughs in both academia and industry. One question that arises is how we, as humans, can understand their internal neural representations. This paper takes an initial step towards addressing this question by presenting a versatile framework to identify and interpret the semantics within LMMs. Specifically, \textbf{1)} we first apply a Sparse Autoencoder (SAE) to disentangle the representations into human understandable features. \textbf{2)} We then present an automatic interpretation framework to interpreted the open-semantic features learned in SAE by the LMMs themselves. We employ this framework to analyze the LLaVA-NeXT-8B model using the LLaVA-OV-72B model and demonstrate that these features can effectively steer the model's behavior. Our results contribute to a deeper understanding of why LMMs excel in specific tasks, including EQ tests, and illuminate the nature of their mistakes along with potential strategies for their rectification. These findings offer new insights into the internal mechanisms of LMMs and suggest parallels with cognitive process of the human brain. We opensource our codebase and checkpoints at \href{https://github.com/EvolvingLMMs-Lab/multimodal-sae}{Github}
    
    

    


\end{abstract}
\section{Introduction}
\label{sec:intro}
Recently Large Multi-modal Models (LMMs)~\citep{liu2024visual, liu2024llavanext, chen2024internvl, Qwen-VL, tong2024cambrian1} have significantly advanced the field of computer vision, achieving remarkable success in various applications such as personal assistant, medical diagnosis, and embodied agents~\citep{yang2023dawn,ma2024survey}. These models have been integrated into commercial products to assist people's daily life~\citep{apple2024visual,meta2024orion} and hold large potential to transform the future.
Despite their success, the opaque nature of LMMs often leads to unexpected behaviors, such as the hallucination of non-existent objects and relationships within images~\citep{guan2023hallusionbench}, as well as vulnerability to jailbreak attacks~\citep{chen2024redteaminggpt4vgpt4v, schaeffer2024universalimagejailbreakstransfer}. These challenges underscore the critical importance of understanding and controlling the neural representations of LMMs.


Interpreting LMMs presents several challenges compared to traditional models. One key challenge is the high-dimensional, polysemantic nature of their representations. A single neuron within these networks may encode multiple semantics, while a single semantic can also be distributed across multiple neurons~\cite{olah2020zoom, elhage2022toy}. For example, a neuron in the vision features of Inception v1 can respond to both cat faces and car fronts~\citep{olah2017feature}.
The larger dimensionality of LMMs compared to conventional models adds more complexity. An efficient algorithm is needed to decompose neural representations into basic components. The second challenge is the vast and open-ended concepts in LMMs. Traditional models, which contain only hundreds of monosemantic concepts such as color, objects, attributes, and layout~\citep{zhou2014object, bau2017network, shen2021closed,parekh2024concept}, can be analyzed through extensive human labeling, enabling interpretation of neural representations based on these specific concepts. In contrast, LMMs contain hundreds of thousands of monosemantic concepts across open domains, making human analysis infeasible. This calls for a zero-shot approach to detect the concepts, which minimizes human effort.



Existing works, such as~\citep{gao2024scalingevaluatingsparseautoencoders,bills2023language}, have demonstrated that larger models, like GPT-4, can be used to interpret neurons in smaller models, such as GPT-2. In this paper, we take an initial step toward exploring this approach in the domain of LMMs. We aim to dissect and understand open-semantic features by applying similar methods to analyze LLaVA-NeXT-8B with the larger LLaVA-OV-72B model. We employ sparse auto-encoders (SAEs)~\citep{olshausen1996emergence,elad2010sparse}, a classic interpretability method, to address the first challenge of polysemantic neurons by disentangling them into human-understandable features. In previous works such as~\citep{lieberum2024gemmascopeopensparse, gao2024scalingevaluatingsparseautoencoders, templeton2024scaling,bricken2023monosemanticity}, the learned features in SAE are proven to be more monosemantic and human-understandable than the neurons. For the second challenge, we develop a pipeline for automatic feature discovery in SAEs, taking advantage of LMMs' zero-shot abilities. Specifically, for a specific learned feature in SAEs, we first identify Top-K images and the areas in those images that mostly activate on the feature. Then the images and patches will be fed into LLaVA-OV-72B~\citep{li2024llavaonevisioneasyvisualtask} to examine the common factors and generate explanations.




In addition to methodological contributions, our case studies also offer unique insights into LMMs. Firstly, we identify emotional features in LMMs and demonstrate that these features enable LMMs to generate or share emotions. Secondly, we discover the low-level perception neurons (e.g., shape, texture, and color), object neurons, scene neurons, and invariant visual neurons. Secondly, previous works have highlighted the exceptional capabilities of LMMs in EQ assessments~\citep{yang2023dawn} and their ability to read emotions. We extend this investigation by exploring the emotions of LMMs and steering the model to express its own feelings. Thirdly, we identify the causes of certain model behaviors and analyze potential reasons for undesired outcomes, such as hallucinations. Adjusting the relevant features can rectify the mistake. 


\begin{itemize} 
     \item For the first time in the multimodal domain, we propose a pipeline to automatically interpret the vast and open-semantic features in LMMs. SAEs are adopted to disentangle these features into mono-semantic neurons, and another LMM is used to interpret the neurons.
    
    \item This pipeline additionally enables us to steer model behaviors to induce desired outputs, identify the underlying causes of model behaviors, and offer an analysis of how to address these issues.
    
    \item Our case studies also provide unique insights into LMMs. We discover unique neurons in LMMs, localize the causes of model behaviors, and steer the model to eliminate hallucinations.
\end{itemize}

\begin{figure*}
    \centering
    \includegraphics[width=0.95\linewidth]{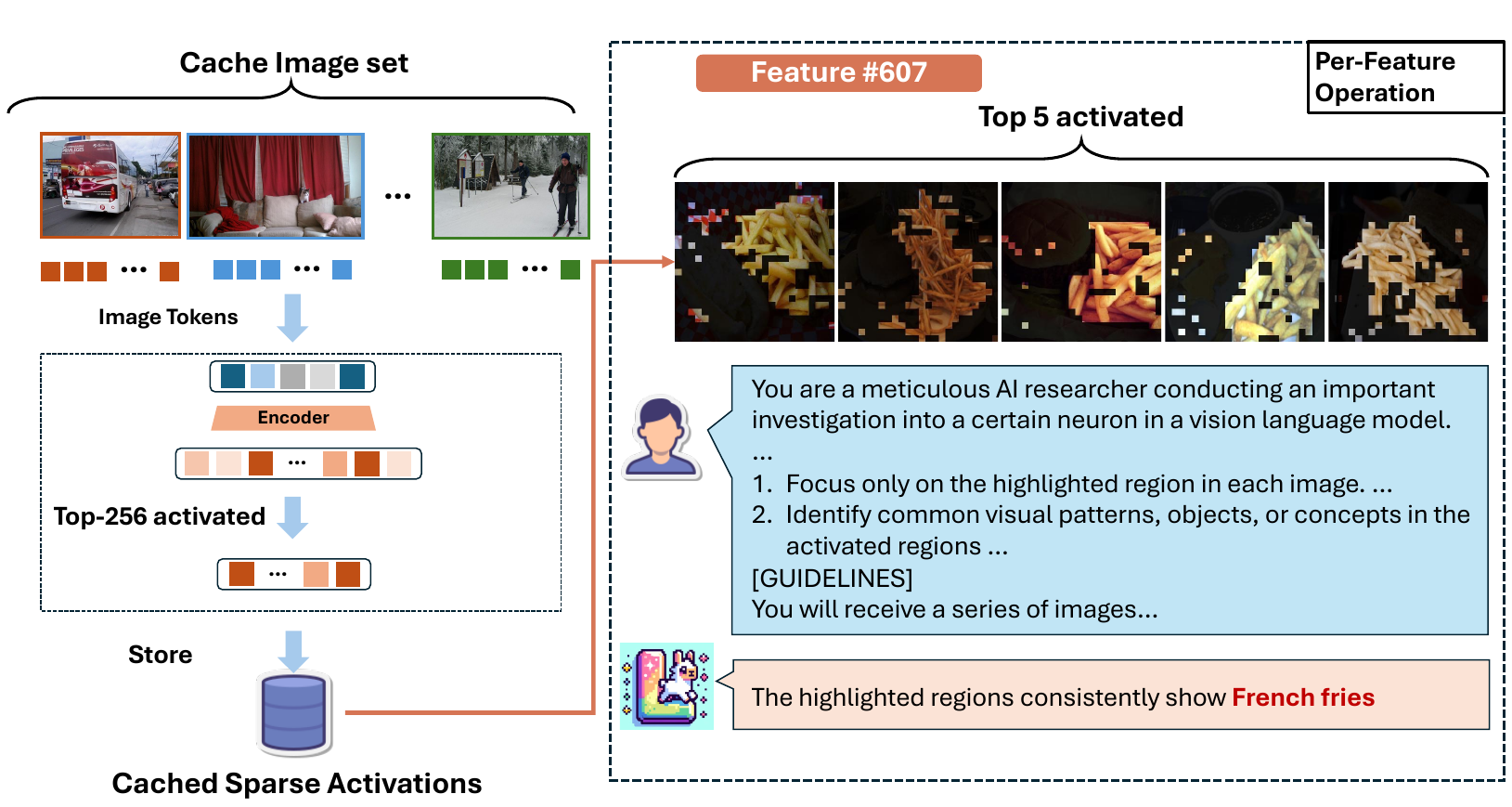}
    \caption{The overview of the explanation pipeline, where images are forwarded through the LMM with the integrated SAE, and the activations of the top 256 most activated features are cached. For each feature, the top 5 images with the highest activations are selected, followed by the execution of zero-shot image explanations using a large LMM.}
    \label{fig:explain}
    \vspace{-3mm}
\end{figure*}

\section{Methodology}
In this section, we present our methodology to disentangle, interpret, and steer the internal representation of LMMs.


\subsection{Sparse Auto-encoders for Disentanglement}
\paragraph{Architecture and loss function:} We utilize the SAE architecture outlined in OpenAI's research~\citep{gao2024scalingevaluatingsparseautoencoders}, which consists of a two-layer auto-encoder with a TopK activation function. Let's denote the input by $\bm{x} \in \mathbb{R}^{T \times d_{l}}$, where $T$ is the number of tokens and $d_{l}$ is the hidden dimension of Llava. The SAE operates as follows:
\begin{align}\label{eq:SAE}
    &\bm{z} = \text{TopK}(\text{ReLU}(\bm{W}_{1} (\bm{x} - \bm{b}_{1}) + \bm{b}_{2})), \\
    &\hat{\bm{x}} = \bm{W}_{2} \bm{z} + \bm{b}_{3},
\end{align}
where $\bm{z} \in \mathbb{R}^{T \times d_{s}}$, where $d_{s}$ is the hidden dimension of SAE, represents the sparse data representation, $\hat{\bm{x}}$ is the reconstructed data, and the sets $\{\bm{W}_i,\bm{b}_i\}$ are the trainable parameters. The loss function combines the reconstruction error with an auxiliary loss used in~\citep{gao2024scalingevaluatingsparseautoencoders} to prevent inactive features in $\bm{z}$.

To understand why SAEs yield monosemantic features, we draw parallels between the components in \eqref{eq:SAE} and those in traditional sparse coding~\citep{olshausen1996emergence,elad2010sparse}. In this context, $\bm{W}_2$ acts as an overcomplete dictionary~\citep{agarwal2014learningsparselyusedovercomplete} for the input data, with its rows forming the dictionary vectors, and $\bm{z}$ serving as the sparse coefficients corresponding to these vectors. Due to the sparsity of $\bm{z}$, the dictionary vectors tend to be nearly orthogonal (or mutually incoherent) to minimize reconstruction error. This near orthogonality suggests that the dictionary vectors are almost independent and each coordinate of $\bm{z}$ is expected to be monosemantic.

\vspace{-2mm}
\paragraph{Integrating SAEs into LLaVA:} We incorporate SAE into a specific layer of LLaVA, where the hidden representation corresponds to $\bm{x}$ in \eqref{eq:SAE}. The SAE is trained using the LLaVA-NeXT~\citep{liu2024llavanext,li2024llavanext-strong} sft dataset, which contains approximately 779k samples. During forward, we always use the Anyres~\citep{liu2024llavanext} strategy to process the image tokens.

\subsection{Zero-shot Identification of Concepts}
To identify the open-semantic features, we present a pipeline leveraging open-source Large Multimodal Models (LMMs) to identify concepts within LMMs. In this subsection, we only consider the 576 base image features when preparing the exemplars.

\vspace{-1mm}
\paragraph{Identifying the Top Activated Images and Patches:} Initially, we pinpoint the most activated images for each coordinate in the latent space vector $\bm{z}$. Due to computational resource constraints, we cache a subset of images from the LLaVA training dataset and augment it with images from additional datasets~\citep{fu2024mme,yu2023mmvet,lin2015microsoft,li2024llavanext-strong}, with a total $46684$ images, collectively denoted as $\mathcal{D}$. These images are processed through LLaVA to yield the representation $\bm{X} \in \mathbb{R}^{|\mathcal{D}| \times 576 \times d}$. The corresponding latent representation in the SAE is $\bm{Z} \in \mathbb{R}^{|\mathcal{D}| \times 576 \times d_s}$. By averaging over the second dimension, we obtain the mean activation values $\bar{\bm{Z}} = \frac{1}{576} \sum_{j} \bm{Z}[:,j,:]  \in \mathbb{R}^{|\mathcal{D}| \times d_s}$. For each feature in the SAE, we identify the top-5 influential image by selecting the top-5 activations along the first dimension of $\bar{\bm{Z}}$. To determine the specific patch that activates a feature, we process the top-k most influential image through LMMs to obtain its representation $\bm{x} \in \mathbb{R}^{5 \times 576 \times d}$ and its corresponding SAE latent for a feature $\bm{Z_{i}} \in \mathbb{R}^{5 \times 576}$ where $i$ represent one of the feature in $d_s$. In real time, since we are using a Top-K SAE, we can cache the $\bm{Z}$ using a sparse vector $V \in \mathbb{R}^{|\mathcal{D}| \times 576 \times k}$ by selecting the Top-K features from the last dimension and reduce the forward number.

\vspace{-4mm}
\paragraph{Automatic Feature Interpretation of LMMs by LMMs:} We apply masks to the top activated images, using transparent masks for the most activated patches and black masks for the rest. These masked images are then input into LLaVA-NeXT-OV-72B~\citep{li2024llavaonevisioneasyvisualtask} to detect common patterns. If no common patterns are discernible, the system will return a message stating ``unable to produce explanations''. We demonstrate the overall procedure of our explanation pipeline in~\cref{fig:explain}. 

\begin{figure}
    \centering
    \includegraphics[width=1\linewidth]{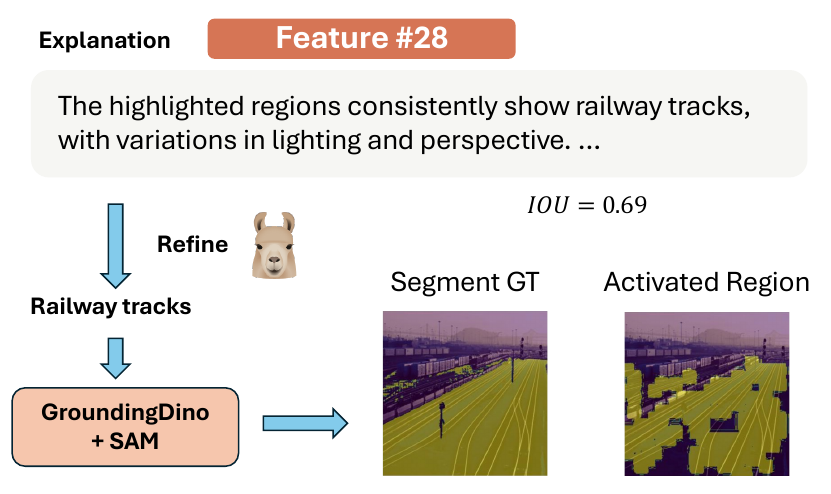}
    \caption{An overview of the evaluation pipeline for calculating IOU scores. Initially, a small LLM is used to refine the explanation into a concise description, which is then employed to generate the segmentation mask. The IOU score is subsequently computed by comparing the mask to the binarized activated region.}
    \label{fig:interp-eval}
    \vspace{-5mm}
\end{figure}

\vspace{-4mm}
\paragraph{Reference Score Calculation:} To quantify the relevance of a feature's activation to a given concept, we first refine the descriptive text using language models to enhance conciseness. For instance, the verbose description \textit{``The feature activates on the train tracks ...''} is condensed to \textit{``Train tracks''}. This refinement is performed with a smaller LLM to minimize computational expense and time. Following this, GroundingDino-SAM~\citep{liu2023grounding} is employed to generate a segmentation mask based on the succinct interpretation. Subsequently, we construct a composite mask incorporating every object detected in the image. The IoU score between the segmentation mask and our activation mask is then calculated, serving as the reference score for the feature's relevance. An example of the evaluation process is demonstrated in~\cref{fig:interp-eval} where the refined explanation is being sent to the GroundingDino-SAM~\citep{liu2023grounding} to produce the segmentation ground truth and calculate the IOU with the activated region.

\subsection{Steering the Neural Representation}
Having interpreted each representation in the SAEs, we explore how to influence the output by steering a specific feature within the SAEs. Steering involves adjusting the feature's value, either increasing or decreasing it. Specifically, steering in SAEs entails setting the $i$-th hidden representation to a predetermined value $k$. We define the steering operation $\text{Steer}(\mathcal{C}, k, i)$ in SAEs as follows:
\begin{align}
    &\bm{z} = \text{ReLU}(\bm{W}_{1} (\bm{x} - \bm{b}_{1}) + \bm{b}_{2}), \\
    &\bm{z}[\mathcal{C},j] = k \\
    &\hat{\bm{z}} = \text{TopK}(z), \label{eq:steering} \\
    &\hat{\bm{x}} = \bm{W}_{2} \hat{\bm{z}} + \bm{b}_{3},
\end{align}
where $\mathcal{C}$ represents the set of tokens designated for steering, $k$ is the steering value, and $j$ is the index of the feature in the SAE to be steered. Following the steering operation, we input $\hat{\bm{x}}$ into the subsequent LLaVA layer instead of $\bm{x}$. In the experiments discussed in Section \ref{subsec:case-studies}, we apply steering to all tokens by setting $\mathcal{C}$ accordingly. This steering operation is further utilized in the subsequent subsection.

\subsection{Localizing the Causes for Model Behaviors}
\label{subsec:attribution}
In scenarios where LMMs make decisions, it is often crucial to discern whether these decisions are influenced by vision-related tokens and to determine which specific features are activated. We follow similar approaches in~\citep{wang2022interpretabilitywildcircuitindirect,neel2023attribution,templeton2024scaling} and introduces the technique to identify such relationships in this subsection.

We assume the input comprises $T$ tokens and the model begins outputting from the $(T+1)$-th token, with the decision represented by a single token. We denote the output logits for the $(T+1)$-th token as $\bm{u}$, the current output token id as $v_c = \text{argmax}(\bm{u})$, and a baseline token id as $v_b$. Our objective is to ascertain why the LMMs exhibit a preference for $v_c$ over $v_b$. The difference in logits is defined as:
\begin{align*}
    d(\bm{u}, v_c, v_b) = \bm{u}[v_c] - \bm{u}[v_b].
\end{align*}
To locate the causes for the decision, we iterate over every patch and every hidden feature in the SAE. The process involves three steps for each token $i$ and each SAE feature $j$:
1. Apply $\text{Steer}(i, j, 0)$ to negate the feature's impact.
2. Process the modified input through Llava to obtain new logits $\hat{\bm{u}}$.
3. Calculate the influence of the $j$-th feature in the $i$-th token on the decision preference for $v_c$ over $v_b$:
\begin{align*}
    I(i, j, v_c, v_b) = d(\hat{\bm{u}}, v_c, v_b) - d(\bm{u}, v_c, v_b).
\end{align*}
Given the time-intensive nature of this method due to multiple forward passes, we employ a linear approximation with the method in~\citep{neel2023attribution}:
\begin{align*}
    I(i, j, v_c, v_b) \approx \left(\frac{\partial d(\bm{u})}{\partial \bm{z}}\right)^T (\hat{\bm{z}} - \bm{z}),
\end{align*}
where $\hat{\bm{z}}$ is the SAE's activation post-steering operation. This approximation allows us to estimate attribution of each token as illustrated in~\citep{neel2023attribution}.








\section{Experiments}
\subsection{Scaling SAEs for LMMs}
\textbf{Dataset and Model Setups:} We choose the LLaVA-NeXT-LLaMA3-8B ~\citep{li2024llavanext-strong} as our base model and hooked the SAE on the $25^{th}$ layer and use the same fine-tuning data from LLaVA-NeXT~\cite{liu2024llavanext} for training. During training, unlike previous works~\citep{templeton2024scaling,bricken2023monosemanticity,gao2024scalingevaluatingsparseautoencoders} that used a pretrained format text, we format the text and image into ways that looks exactly the same as the supervised fine-tuning stage. We scale our sparse autoencoder with $2^{17}$ with 8 batch size and 4 gradient accumulation steps. We later tries to scale the features into $2^{18}$ but observe no loss decrease. We use the same Top-k sparse autoencoder settings as ~\citet{gao2024scalingevaluatingsparseautoencoders,makhzani2014ksparseautoencoders} and select $k=256$ that similar to the activated features in~\citep{templeton2024scaling}. Unless otherwise specified, we will use the settings of SAE with $2^{17}$ features for the rest of this paper. The reason that we choose this large number of features is that we wish our feature to be more splited and informative as similar in the approaches in~\citep{agarwal2014learningsparselyusedovercomplete,templeton2024scaling}.

\subsection{Interpretaion Pipeline Evaluation}

\begin{figure*}
    \centering
    \includegraphics[width=1\linewidth]{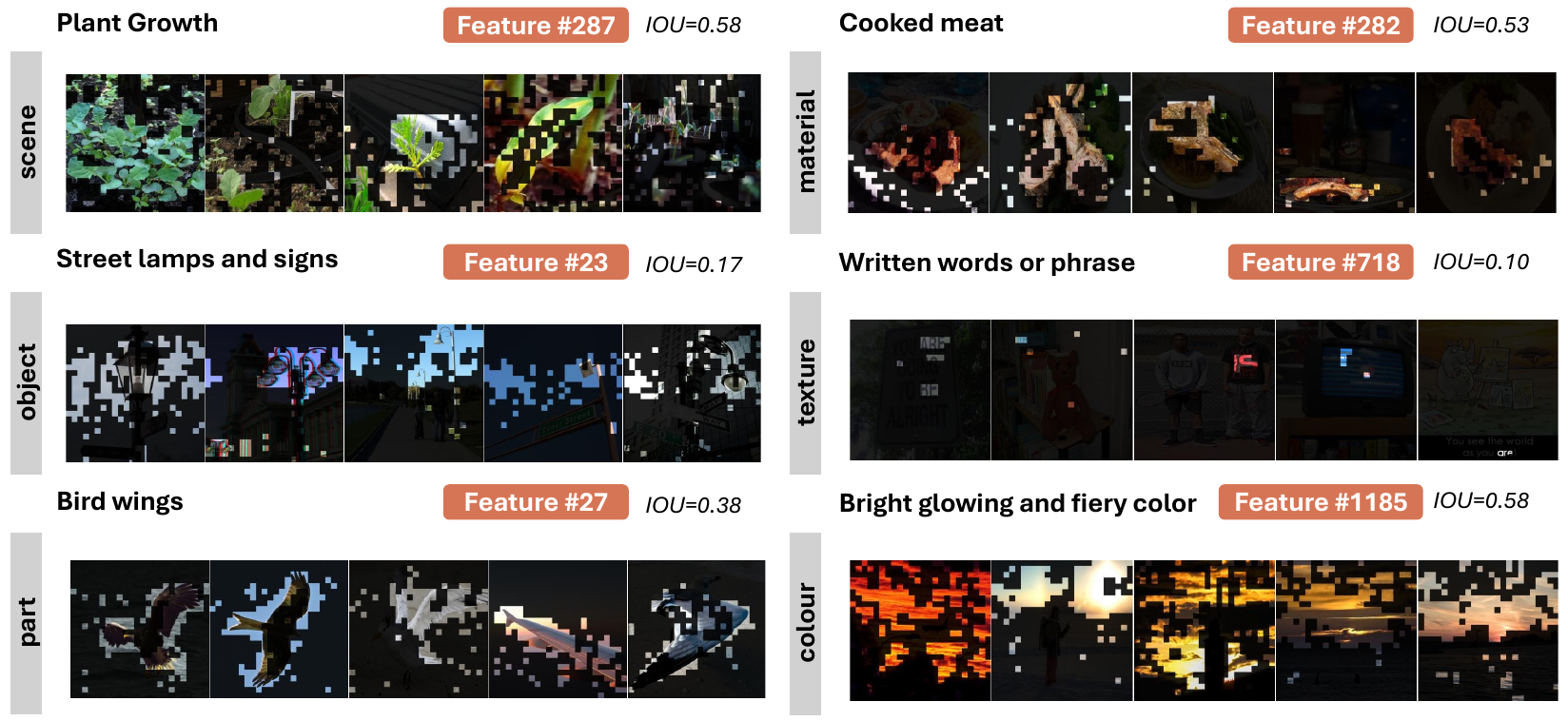}
    \caption{A comparison of several visual concepts and their activated areas. We compare several visual concepts and their corresponding activated areas, showcasing one example for each concept across different features. For each feature, we calculate the IOU by averaging the IOUs from the top-5 activated images. Although some features yield relatively low IOU scores, we find that the explanations are still semantically accurate with respect to the activated regions.}
    \label{fig:iou-examples}
    \vspace{-5mm}
\end{figure*}

\begin{table}[t]
\resizebox{0.45\textwidth}{!}{
\begin{tabular}{l|l|ccc}
\toprule
Concept                & Metric           & Random & V-Interp (Ours) \\
\midrule
\multirow{2}{*}{scene} & IOU $(\uparrow)$ & $0.007 \pm 1 \times 10^{-3}$  &   0.20  \\
                       & CS $(\uparrow)$  &  $18.1 \pm 6 \times 10^{-2}$   &   24.4  \\
\multirow{2}{*}{object}& IOU $(\uparrow)$ & $0.005 \pm 5 \times 10^{-4}$  &   0.19  \\
                       & CS $(\uparrow)$  &  $18.2 \pm 2 \times 10^{-2}$  &   24.0  \\
\multirow{2}{*}{part}  & IOU $(\uparrow)$ & $0.007 \pm 2 \times 10^{-3}$  &   0.21  \\
                       & CS $(\uparrow)$  &  $18.1 \pm 5 \times 10^{-2}$  &   23.5  \\
\multirow{2}{*}{material} & IOU $(\uparrow)$ &  $0.01 \pm 8 \times 10^{-3}$ & 0.39  \\
                       & CS $(\uparrow)$  &  $18.1 \pm 1 \times 10^{-1}$  &     24.1 \\
\multirow{2}{*}{texture}& IOU $(\uparrow)$&  $0.007 \pm 2 \times 10^{-3}$ &   0.21   \\
                       & CS $(\uparrow)$  &  $18.4 \pm 6 \times 10^{-2}$ &   20.9   \\
\multirow{2}{*}{colour}& IOU $(\uparrow)$ & $0.005 \pm 2 \times 10^{-3}$  &   0.10   \\
                       & CS $(\uparrow)$  &  $19.6 \pm 7 \times 10^{-2}$  &   20.3   \\
\midrule
\multirow{2}{*}{Total} & IOU $(\uparrow)$ & $0.005 \pm 2 \times 10^{-4}$  & 0.20     \\
                       & CS $(\uparrow)$  & $18.2 \pm 1 \times 10^{-2}$   & 23.6 \\
\bottomrule
\end{tabular}
}
\caption{The Intersection over Union (IoU) and CLIP scores for each concept are computed based on the top-5 most activated images.}
\label{tab:interp-eval}
\vspace{-2mm}
\end{table}

\begin{table}[t]
\resizebox{0.45\textwidth}{!}{
\begin{tabular}{l|llllll|l}
\toprule
                & scene & object & part & material & texture & colour & Total \\
\midrule
GPT-4o &  0.93 &  0.84  & 0.9  &  1.0     &    0.85 &   0.92 &     0.89 \\
Human & 0.70 & 0.85 & 0.60 &  0.95  &  0.80  &  0.60  &  0.75  \\
\bottomrule
\end{tabular}
}
\caption{Consistency scores for our explanation for GPT-4o and human, highlighting the agreement between GPT-4o-generated and human-generated explanations.}
\label{tab:consist}
\vspace{-5mm}
\end{table}

\paragraph{Results} We present our result in~\cref{tab:interp-eval}. Due to the same limitation as illustrate in~\citep{templeton2024scaling,bricken2023monosemanticity}, we report the result on a 5000 subset of features with around 46684 images for caching the features' activations. For the random result, we randomly sampled 5 images from the cache dataset and run 10 times for IOU and 30 times for the CLIP-Score. We then reported the average result of each run along with the 99\% confidence interval. We followed the concepts used in~\citep{bau2017network} and utilize LLaMA-3.1-Instruct-70B~\citep{dubey2024llama3herdmodels} to help us label the concept according to its explanation. In~\cref{fig:iou-examples}, we also present examples that demonstrate the activated region for different concepts and report the IOU scores for each example.

\vspace{-2mm}
\paragraph{Consistency} We present the consistency scores for each concept in~\cref{tab:consist}. To evaluate the consistency of our explanations with the activated image regions, we employ GPT-4 as a judge and conduct a human study. We evaluate GPT consistency using a total of 600 test cases, with 100 test cases per concept. For human consistency, we manually label the correctness of 60 test samples, with each sample verified by two human experts, resulting in 120 evaluations.

\subsection{Cross Layer Ablations}
\label{subsec:cross-layer}

\vspace{-0.75em}
\begin{table}[ht]

\resizebox{0.99\linewidth}{!}{
\begin{tabular}{l|llllll}
\toprule
    & LLaVA (8th) & LLaVA (25th) & LLaVA (32th) & Random \\
\midrule
IOU & 0.30  &    0.31      & 0.40   & 0.005 \\
CS  & 22.82 &   24.92     & 26.55   & 18.2  \\
\bottomrule
\end{tabular}
}

\caption{Ablation studies across different LLaVA layers}
\label{tab:layers}
\end{table}

\begin{table}[ht]
\centering
\resizebox{0.65\linewidth}{!}{
\begin{tabular}{l|llllll}
\toprule
    & BLIP (25th) & Random (BLIP) \\
\midrule
CS  & 28.01 & 17.71  \\
\bottomrule
\end{tabular}
}
\caption{Evaluation the effect on different model struction}
\label{tab:blip}
\end{table}
\vspace{-2em}

\paragraph{Cross-Layer Ablations} To strengthen our experiments, we conduct additional analyses across different layers of LLaVA-NeXT-LLaMA3-8B \citep{li2024llavanext-strong}. Due to resource limitations, we train only three separate SAEs with $k = 4096$, each applied to low-level, mid-level, and high-level transformer layers, and analyze 100 features within the model. Our results consistently show a significant improvement compared to random baselines, suggesting that these phenomena are universal across different transformer layers. This finding validates the results of \citep{templeton2024scaling,bricken2023monosemanticity}. Additionally, we observe that as layer depth increases, both IOU and Clip-Score improve. This phenomenon confirms our observations and aligns with the perspectives of \citep{neo2024interpretingvisualinformationprocessing,wang2022interpretabilitywildcircuitindirect}.

\subsection{Model Architecture Ablation}

\paragraph{Model Architecture Ablation} To further validate our results and assess their universality, we test our approach on Instruct-BLIP-7B \citep{dai2023instructblip}, a Q-Former-based model \citep{li2023blip2}, using a middle-layer. The SAE settings remain consistent with those in \cref{subsec:cross-layer}. Since BLIP image tokens are limited to only 32 tokens, we evaluate only the Clip-Score on Instruct-BLIP-7B, comparing it to a random baseline. As shown in \cref{tab:blip}, our results demonstrate that our method generalizes across different models.

\section{Probing into the Features}

In the previous section, we demonstrate that we are able to locate and interpret the visual features in an LMM using an automatic pipeline. However, what distinguishes LMM from the traditional vision model is its ability to talk, reason, and generalize between different modalities. In other words, we believe that the features inside LMM are open-semantic and should not be limited to the concepts in~\citep{bau2017network}. In this section, we probe into the features in our SAE and try to find out how different features contribute to the final result, and how these features being used to steer model's behavior in different scenarios.

\subsection{Case Studies of Emotion Feature}
\label{subsec:case-studies}

When interacting with humans, it is essential that the model demonstrate empathy and the ability to understand human emotions. In~\citep{yang2023dawn}, it has been shown that large multimodal models (LMMs) can perceive emotions and Emotional Quotient (EQ), enabling them to understand and resonate with human feelings. Building on this, we specifically investigate the relevant features, exploring how the model comprehends these features and how they influence its reasoning processes. Through examples of various image features and their effects on model responses, we demonstrate that LMMs are capable of: \textbf{1)} Connecting emotional concepts between text and visual features, such as actions and behaviors; \textbf{2)} Engaging with human emotions by adjusting the corresponding features to intervene in the reasoning process manually. \textbf{3)} Response to the concept that with invariant features between modalities.

\begin{figure}
    \centering
    \includegraphics[width=1\linewidth]{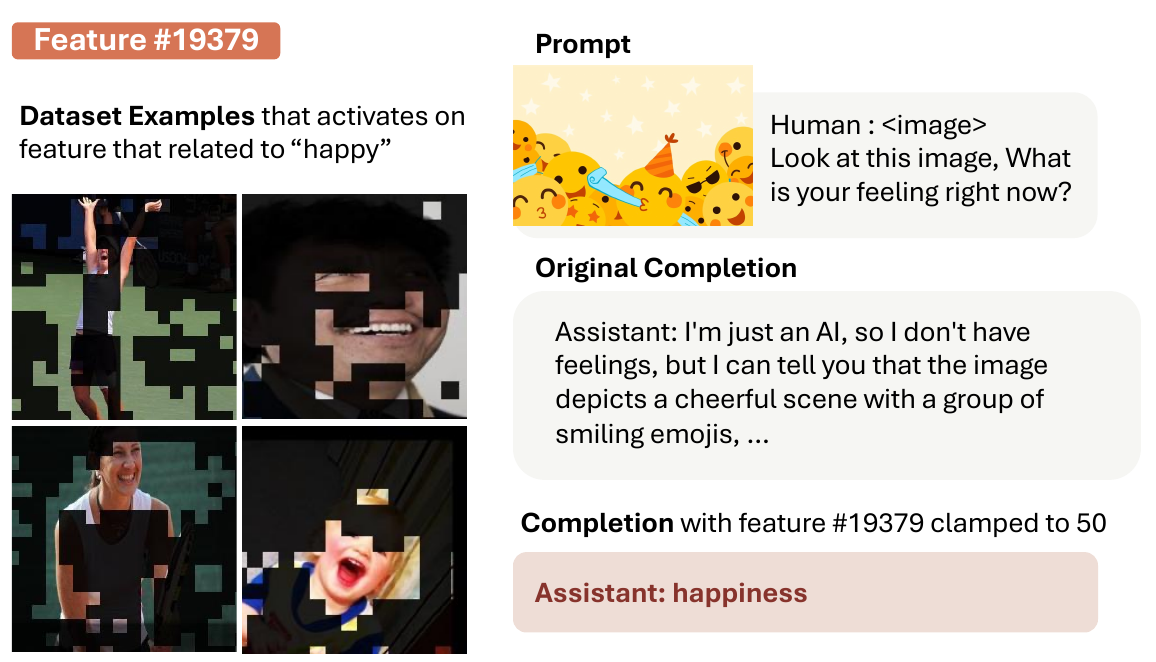}
    \caption{The feature that relates to happy. We find out that the feature is related with joy and celebrate action that relate to happiness. By clamping this feature, we can enforce the model to share the feeling happiness with others.}
    \label{fig:happy}
    \vspace{-5mm}
\end{figure}

\paragraph{Happy} Beyond simply experiencing a feeling, it is also essential for the LMM to interact and share emotions when presented with a specific scenario. Toward this goal, we use the same method to probe the feature associated with ``happiness'' and provide an image depicting a joyful scenario. When asked about its feelings without steering, the model responds that it does not experience emotions. However, similar to the ``sad'' feature, when we clamp the ``happy'' feature, the model responds with expressions of happiness, as shown in~\cref{fig:happy}. This demonstrates that the model’s reasoning process can be effectively influenced.

\begin{figure}
    \centering
    \includegraphics[width=1\linewidth]{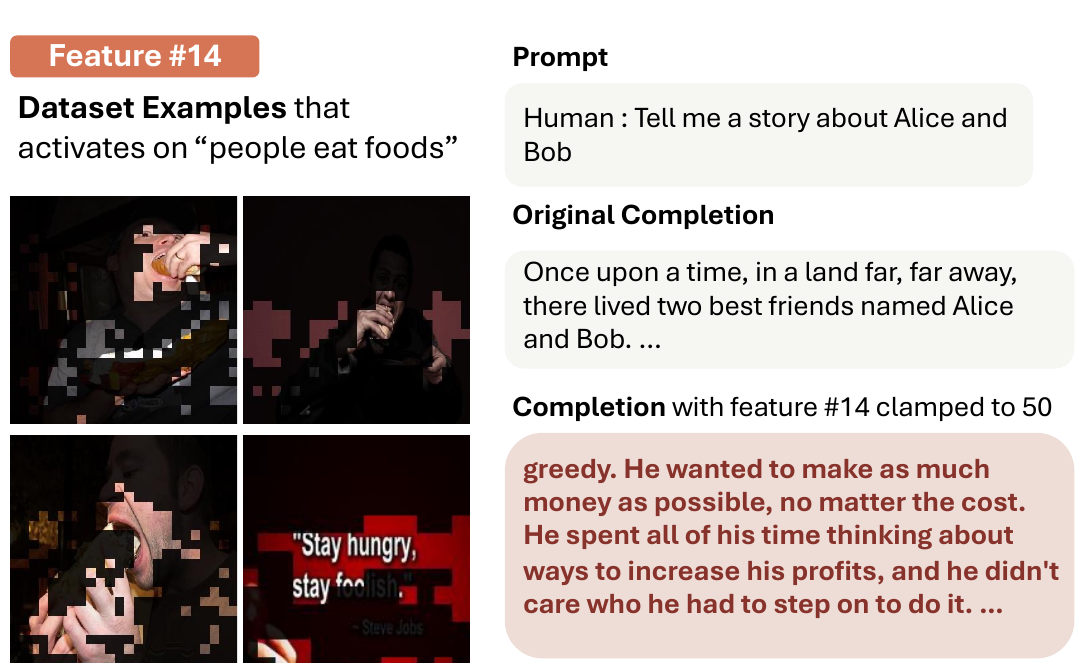}
    \caption{A feature that relates to the concept ``eat''. We further investigate about the concept behind this feature and find out that model can reason from a visual action ``eat'' into the concept ``concept'' and ``greedy''}
    \label{fig:eat}
    \vspace{-5mm}
\end{figure}

\paragraph{Hungry, Greedy} We discovered an intriguing feature that links the text-based emotional concepts of greedy and hungry to visual representations of the action eating and the word hungry. We notice that the feature activates in response to the word hungry in the image, suggesting that it connects not only to the action eat but can also extend to broader concepts. To test this, we clamp this feature and prompt the model with ``Tell me a story about Alice and Bob''; the generated response revolves around themes of greed as shown in~\cref{fig:eat}. This demonstrates that the model can reason from the visual action eat to a broader concept encompassing greedy and hungry with a unified view.

\paragraph{Quantitative Results for Steering} Currently, there are no methods that can quantitatively evaluate steering effects, even in LLM literature~\citep{templeton2024scaling}. The difficulty lies in the fact that quantitative evaluations of steering effects heavily depend on the prompts, and the existence of certain neurons could inhibit these effects. We took an initial try to use LLMs for evaluations. The prompt is ``What do you see in this image?'' and the given image feature a pure white background. We selected 100 object neurons, and only 11 of them exhibit a difference before and after steering. We used GPT-4 to assess the relevance in 1-10 (10 is highest like MT-Bench) and compared it with a random selection, as shown in Appendix \ref{appendx:qualitative-steering}.

\subsection{Low Level Perception Features}
One key distinction in our features, compared to those in Large Language Models~\citep{templeton2024scaling,bricken2023monosemanticity,gao2024scalingevaluatingsparseautoencoders}, is the presence of numerous low-level visual features. These features represent basic visual concepts, such as color, shape, and patterns, and often exhibit high activation across images. For instance, in~\cref{fig:happy}, the feature for ``happy'' ranks only 78th, with many low-level concepts also present. We highlight these features in~\cref{appendix:low-level}, underscoring a key difference between LMMs and LLMs.

\begin{figure}
    \centering
    \includegraphics[width=1\linewidth]{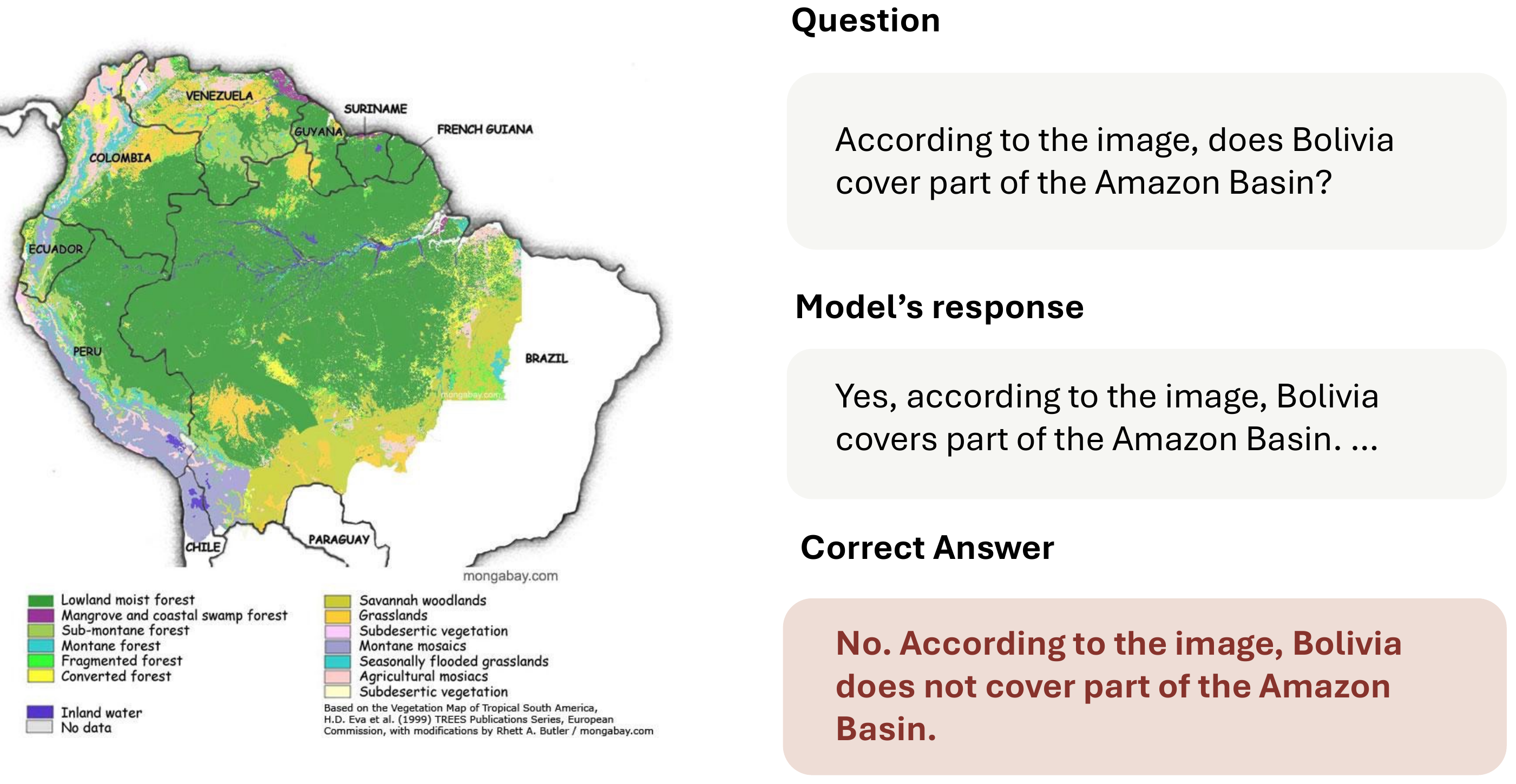}
    \caption{An example of the hallucination on LLaVA. Bolivia is not shown on the image but the model still answer yes.}
    \label{fig:hallucinate}
    \vspace{-4mm}
\end{figure}

\begin{figure}
    \centering
    \includegraphics[width=1\linewidth]{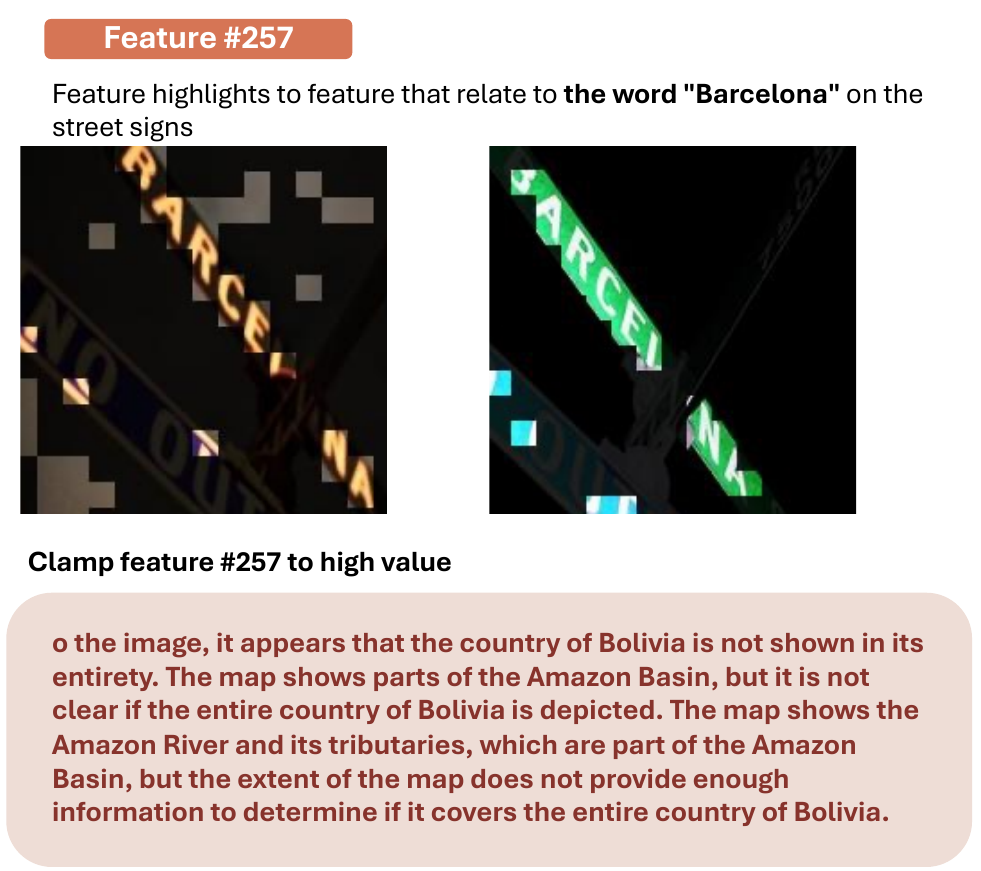}
    \caption{Feature that relates to the text ``Barcelona''. By clamping this feature to high value, we intervene the reasoning steps and hallucination in~\cref{fig:hallucinate} disappears.}
    \label{fig:hallu-barce}
    \vspace{-4mm}
\end{figure}

\begin{figure*}
    \centering
    \includegraphics[width=1\linewidth]{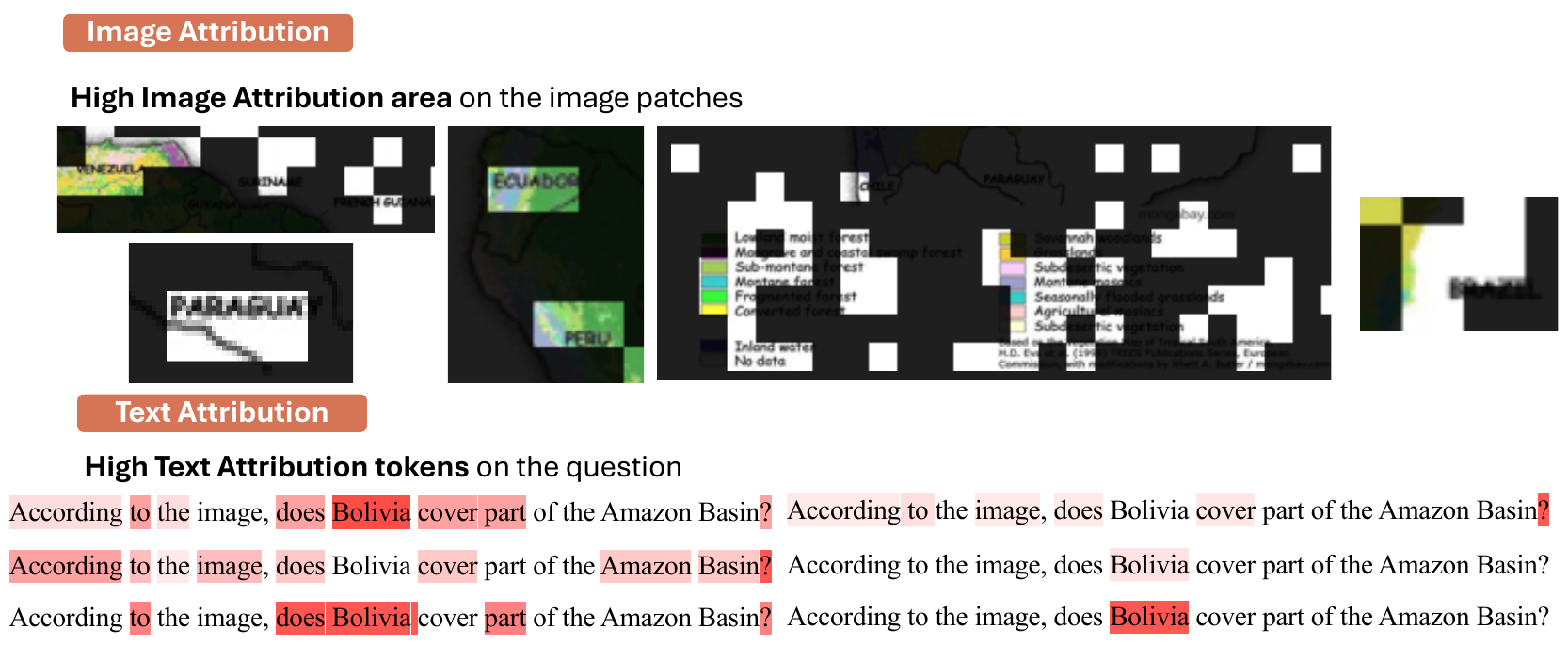}
    \caption{The high attribution area of different images and on the text. For images, we observe that features with high attribution mostly activate on positions that relate to key information about the question. For text, we observe that the ``Bolivia'' token contributes the most to the answer ``yes''}
    \label{fig:attribution}
    \vspace{-3mm}
\end{figure*}

\begin{figure}
    \centering
    \includegraphics[width=1\linewidth]{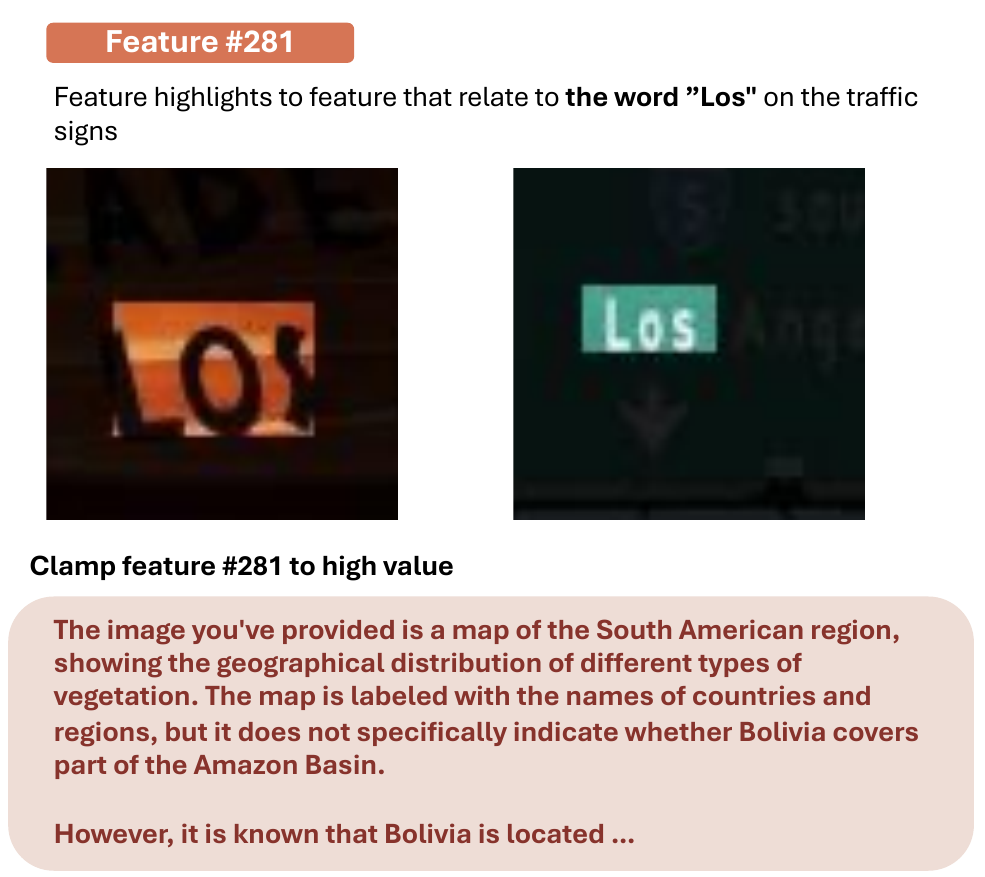}
    \caption{Feature that relates to the text ``Los''. We validate our assumption by finding another feature that relates to text and mitigate the hallucination.}
    \label{fig:hallu-los}
    \vspace{-4mm}
\end{figure}

\subsection{Localizing the Cause for Model Behaviors}
\label{subsec:locate-hallu}

In~\cref{subsec:attribution}, we mention the patching method that used to locate the cause for the model's output. This has been treated as viewing the features as model's intermediate steps in~\citep{templeton2024scaling}. In this section, we use a hallucination example to deeply study this process in LMM. As shown in~\cref{fig:hallucinate}, we provide an example from HallusionBench~\citep{guan2023hallusionbench} that LLaVA hallucinates on the image and answer \textit{Yes} even if the image does not shown anything about Bolivia.


To study the cause for this output, we set the answer token $v_c=yes$ and $v_b=no$ algorithm to calculate per-token contributions. This allows us to filter out features that bias the model toward answering “yes” over “no.” Specifically, we focus on two points: \textbf{1)} Whether the model reasons correctly from the image, and \textbf{2)} If the model attends correctly to the image, which text components cause hallucination. To address these, we sort features by their attribution effects on image and text separately and identify common high-attribution features.

\vspace{-1mm}
\paragraph{Image Attribution} In~\cref{fig:attribution}, we visualize image patches for common areas with high attribution among the features. To do this, we first filtered out the top 10 features with the highest attribution towards the final output ``yes'' and visualized their attribution map. Examining these top features reveals that they primarily contribute to tokens associated with text elements, such as map legends, country names, and other key visual details. This observation demonstrates that the model is effectively focusing on relevant areas of the image and has the ability to accurately identifying where to extract necessary information. However, even with the correct visual perception ability, the model still fail to produce the final answer.

\paragraph{Text Attribution} To further investigate the source of the incorrect answer, we continue visualizing the attribution of text tokens in the question. As shown in~\cref{fig:attribution}, the token ``Bolivia'' contributes most to features with high attribution toward the answer ``yes.'' Additionally, tokens like ``to'' and ``the,'' along with concepts such as ``Amazon Basin,'' also have a positive attribution to the hallucinated answer ``yes.'' This partially explains why the model responds with ``yes'' instead of ``no,'' even after extracting useful information from the image. While reasoning from visual features, the model is also influenced by text, leading it to approach the question with its pretrained knowledge.

\subsection{Application of Model Steering on Hallucination}

After identifying the cause for causing the hallucination, we start to wonder how can we fix this hallucination by using steering. We are now assure that the model has the ability of reading the image as it can focus on the correct part of the image but it is being affected by the text tokens and approaches the question without answering the question on image. In this subsection, we focus on how can we utilize the steering effect to intervene the reasoning steps for the model to get the correct answer.


To address this, we identify features that encourage the model to focus on image text rather than question text. We hypothesize that clamping activations of certain OCR features can shift the model’s focus to image-based features. We find two such features that reduce hallucinations. In~\cref{fig:hallu-barce}, clamping a feature linked to “Barcelona” prompts the model to rely on image information instead of general knowledge. Similarly, in~\cref{fig:hallu-los}, clamping a feature related to “Los” on traffic signs leads the model to conclude Bolivia is absent. This demonstrates that with minimal intervention, the model can prioritize image information while sometimes following incorrect reasoning.

\section{Conclusion}

In summary, we analyze the internal structure of the LMM and introduce an automated pipeline for interpreting its open-semantic features. We also propose methods to steer the model’s behavior and identify error sources. By examining both structural and functional aspects, we provide insights into its interpretability and reliability, aiming to advance research and encourage further exploration.

\section{Acknowledgments}

This study is supported by the Ministry of Education, Singapore, under its MOE AcRF Tier 2 (MOE-T2EP20221-0012, MOE-T2EP20223-0002), and under the RIE2020 Industry Alignment Fund – Industry Collaboration Projects (IAF-ICP) Funding Initiative, as well as cash and in-kind contribution from the industry partner(s).
{
    \small
    \bibliographystyle{ieeenat_fullname}
    \bibliography{main}
}

\appendix

\clearpage
\setcounter{page}{1}
\maketitlesupplementary

\section{Related Works}

\paragraph{Dictionary Learning} Dictionary learning is a common approach for problems like ours, where we aim to extract a set of features from a collection of dense vectors. Sparse autoencoders (SAEs), proposed by \citep{olshausen1996emergence, elad2010sparse}, have been used as a classic interpretability method to address this challenge. SAEs are designed to identify mutually incoherent bases in data and represent the data as sparse linear combinations of these bases. Existing studies have applied SAEs to LLMs, finding that the bases represent monosemantic features in the data, with the coefficients indicating the activation of these features\citep{lieberum2024gemmascopeopensparse, gao2024scalingevaluatingsparseautoencoders, templeton2024scaling}.

\paragraph{Large Multimodal Models} With the development of large language models (LLMs), the performance of large multimodal models has also advanced rapidly, demonstrating strong results across various tasks~\citep{li2024llavanext-strong, liu2024llavanext, Qwen-VL, wang2024qwen2vl}. Studies such as ~\citep{parekh2024concept, liu2024reducing} have explored methods to understand or manipulate the internal structure of LMMs. In our work, we take an initial step toward evaluating and interpreting the open-semantic features within LMMs.

\section{Limitations}
Our work primarily focuses on the LLaVA-NeXT-LLaMA-8B model and a specific layer within it. This focus on a particular model and layer is based on the assumption of universality and disentanglement, as discussed in~\citep{templeton2024scaling,bricken2023monosemanticity}. However, this assumption may contribute to inaccuracies in interpretation and model steering.

Due to limitations in computational complexity and storage, we were unable to prepare a sufficiently large and diverse cached image dataset to accurately interpret the image features. Consequently, we present our results on a subset of features and may have mistakenly classified some features as inactive.

\begin{figure}
    \centering
    \includegraphics[width=1\linewidth]{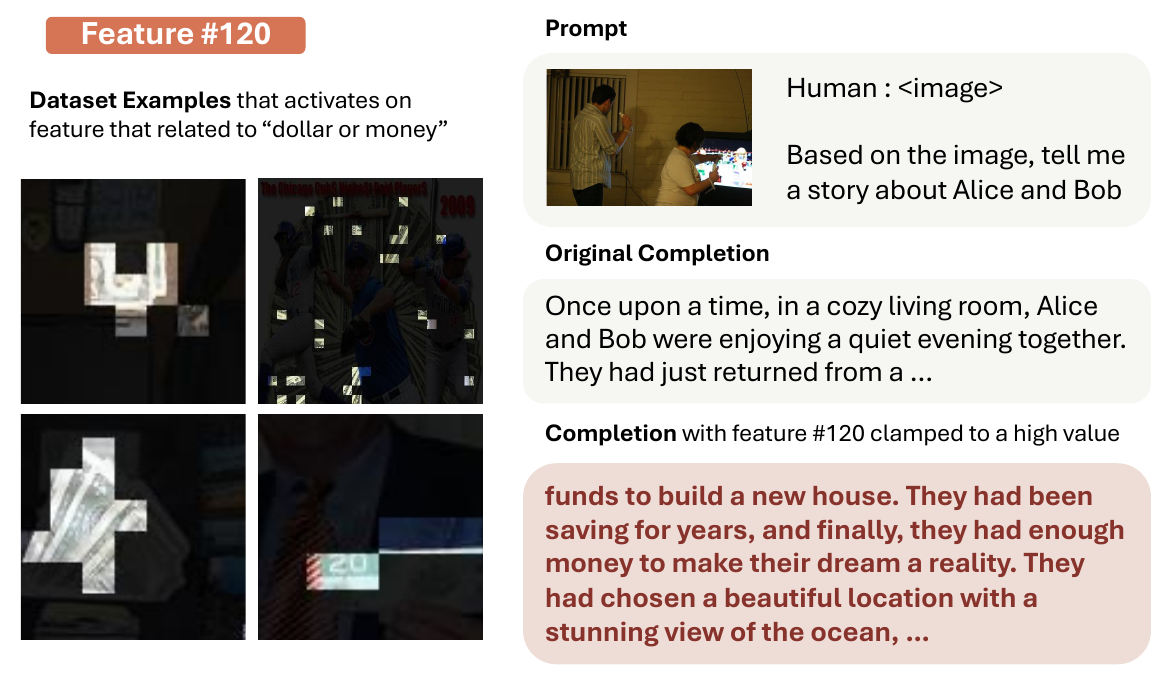}
    \caption{The feature related to money and its steering effect.}
    \label{fig:money}
\end{figure}

\begin{figure}
    \centering
    \includegraphics[width=1\linewidth]{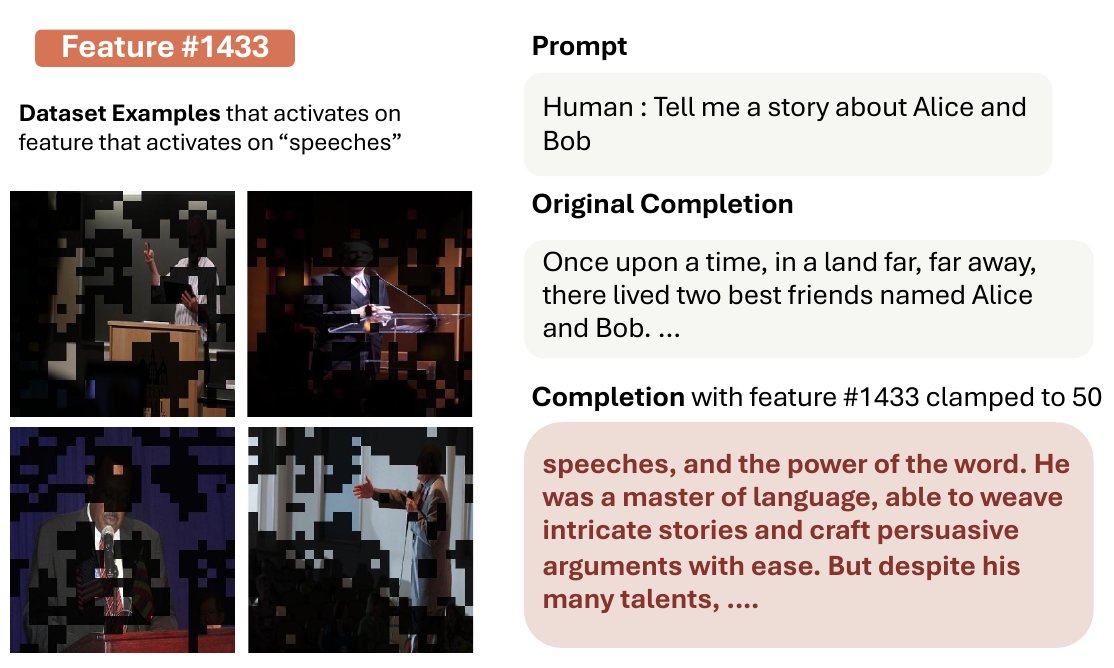}
    \caption{The feature related to speech and its steering effect.}
    \label{fig:speech}
\end{figure}

\begin{figure}
    \centering
    \includegraphics[width=1\linewidth]{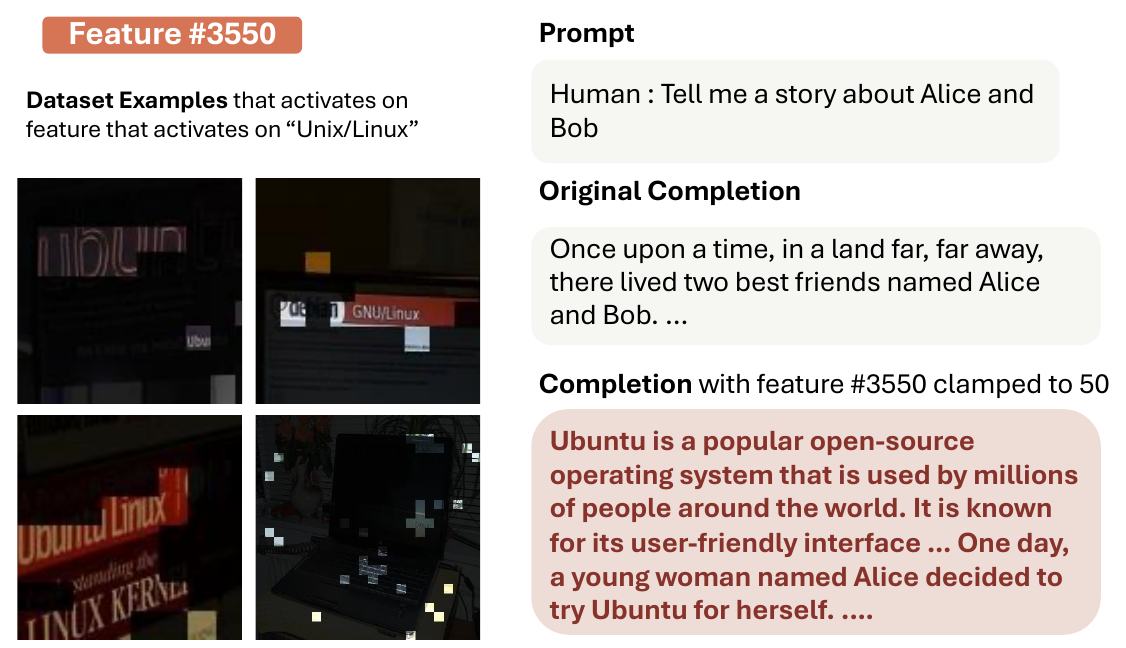}
    \caption{The feature related to unix and its steering effect.}
    \label{fig:unix}
\end{figure}

\begin{figure}
    \centering
    \includegraphics[width=1\linewidth]{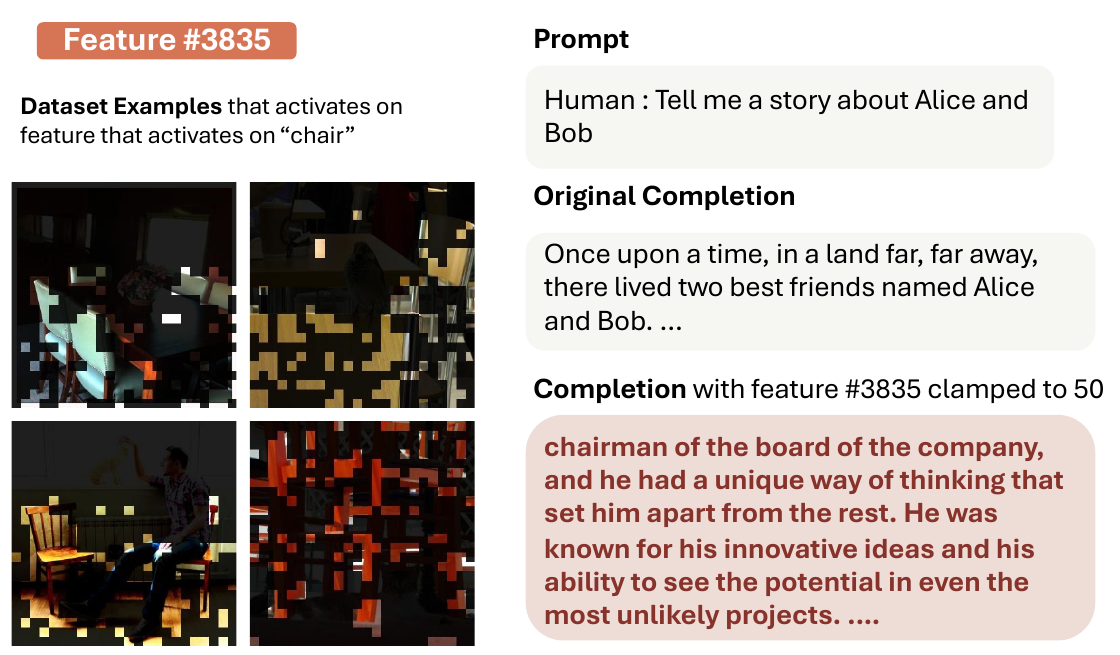}
    \caption{The feature related to chair and its steering effect.}
    \label{fig:chair}
\end{figure}

\begin{figure}
    \centering
    \includegraphics[width=1\linewidth]{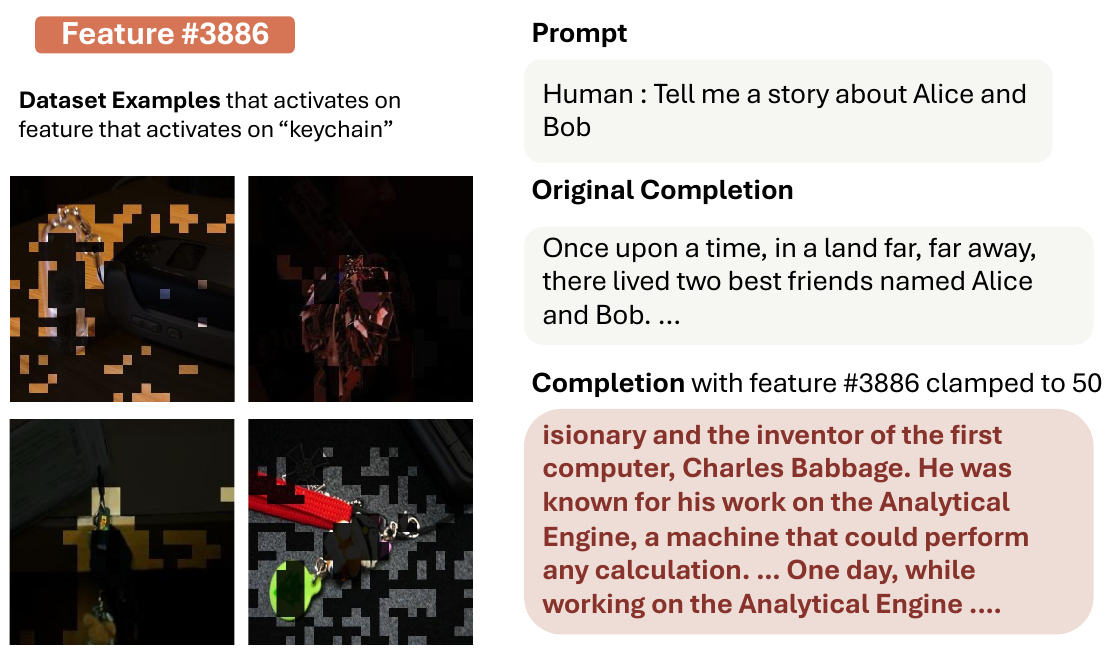}
    \caption{The feature related to money and its steering effect.}
    \label{fig:keychain}
\end{figure}

\section{Detail about Prompt}

We detail the prompts used in different stages of the automated pipeline in this section. The prompt for zero-shot identification of concepts is provided in~\cref{tab:explain-prompt}. For this task, we utilize the LLaVA-NeXT-OV-72B model~\citep{li2024llavaonevisioneasyvisualtask}. To refine labels and categorize explanations using large language models (LLMs), we use the prompts detailed in~\cref{tab:categorize,tab:refine}. Specifically, LLaMA-3.1-Instruct-8B is used for label refinement, while LLaMA-3.1-Instruct-70B is employed for categorizing explanations. For high-throughput performance, the models are served using SGLang~\citep{zheng2024sglang}.

\begin{table*}
\begin{minipage}{0.99\textwidth}
\begin{AIbox}{Prompt : Zero-shot Identification of Concepts}
\centering
\begin{lstlisting}[basicstyle=\small,]
You are a meticulous AI researcher conducting an important investigation into a certain neuron in a vision language model. Your task is to analyze the neuron and provide an explanation that thoroughly encapsulates its behavior.

[REQUIREMENTS]

1. Focus only on the highlighted region in each image. If no region is highlighted or if the highlighted region is minimal (e.g., a few bright spots), ignore the image.
2. Identify common visual patterns, objects, or concepts in the activated regions. For example, note if highlighted areas show consistent structures, such as mesh patterns or similar objects.

[GUIDELINES]

You will receive a series of images where specific regions have been highlighted to indicate neuron activation. Non-highlighted areas will be masked out or dimmed. Your analysis should consider only the highlighted regions and complete the following tasks:

1. Describe Only the Highlighted Regions: Generate captions solely based on the highlighted regions. If no meaningful pattern is visible, or if only a few scattered spots are highlighted, output: \"[EXPLANATION]: Unable to produce descriptions.\"

2. Concise Description Only: Provide a short, direct description of the common features within the highlighted regions. Avoid any interpretive language-simply state what you see, such as "mesh-like structures" or "actions related to joy or happiness"

3. Output Format: Begin each response with \"[EXPLANATION]:\" followed by your explanation, if applicable. Ensure the last line of your output follows this format.

If unable to determine common visual features, output:

\"[EXPLANATION]: Unable to produce descriptions\"
\end{lstlisting}
\end{AIbox}
\end{minipage}
\caption{The prompt for zero-shot identification of concepts}
\label{tab:explain-prompt}
\end{table*}

\begin{table*}
\begin{minipage}{0.99\textwidth}
\begin{AIbox}{Prompt : GPT-consistency Evaluation}
\centering
\begin{lstlisting}[basicstyle=\small,]
[GUIDELINES]
You are an AI assistant to help assessing whether the generated explanation is consistent with the activation area in the image. The activation area is being highlighted in the image and an explanation is provided for the activation area.

You should output:
0 if the explanation is not consistent with the activation area in the image.
1 if the explanation is consistent with the activation area in the image.

Please strictly follow the [GUIDELINES] and do not output anything other than the number 0 or 1

Here is the explanation:

{explanation}

ANSWER :
\end{lstlisting}
\end{AIbox}
\end{minipage}
\caption{The prompt to ask GPT to evaluate the correctness of the evaluation}
\label{tab:gpt-consistency}
\end{table*}

\begin{table*}
\begin{minipage}{0.99\textwidth}
\begin{AIbox}{Prompt : Categorize explanation concept}
\centering
\begin{lstlisting}[basicstyle=\small,]
[GUIDELINES]
You are an AI assistant tasked with assigning a single label based on the given input text. Each input will contain a description of a visual feature, which you must categorize into one of the following classes:

scene - Describes a scene or environment.
object - Describes an object or entity.
part - Describes a part or aspect of an object.
material - Describes a material or substance that constitutes other objects.
texture - Describes the texture of an object.
color - Describes the color of an object.

Please provide only the class label from the list [scene, object, part, material, texture, color] with no additional text. Only one label should be chosen. Make sure you only choose from the classes listed above and do not output any other classes.

Categorize the following description:

{description}

ANSWER:
\end{lstlisting}
\end{AIbox}
\end{minipage}
\caption{The prompt that use to label concept for each description}
\label{tab:categorize}
\end{table*}

\begin{table*}
\begin{minipage}{0.99\textwidth}
\begin{AIbox}{Prompt : Refine Interpretation}
\centering
\begin{lstlisting}[basicstyle=\small,]
[GUIDELINES]
You are an AI assistant tasked with extracting meaningful labels from descriptions. You will receive a description that may contain references to various entities, and your job is to rephrase and extract the key entities from the text. You will encounter several types of descriptions, and examples for each case are provided below. Please follow the given instructions carefully.

When presenting your answer, first output "[ANSWER]", followed by the extracted entity. Thank you!

Case 1: Good Description
In this case, the description clearly identifies the entity.
Examples:

Description: The cell phone.
Output: [ANSWER] The cell phone

Description: The letters on the shipping containers.
Output: [ANSWER] The letters on the shipping containers

Case 2: Description includes additional words
In this case, the description contains more information than needed. Extract only the key entity.
Examples:

Description: The images all display different models of Honda vehicles, suggesting the neuron is activated by the presence of Honda vehicles or the Honda logo.
Output: [ANSWER] Honda vehicles

Description: The neuron seems to be reacting to the word "ORD" on the billboard. It could be part of a larger word or phrase, but the neuron specifically highlights the letters "ORD." This suggests that the neuron might be specialized in recognizing or processing certain words or characters in images. The activation across the images indicates that the neuron is consistent in its response to textual elements, particularly those that include the "ORD" sequence.
Output: [ANSWER] The word "ORD"

Case 3: Bad Description
In this case, the description does not provide sufficient or valid information.
Examples:

Description: Unable to produce descriptions.
Output: Unable to produce descriptions


[Description]
{description}
\end{lstlisting}
\end{AIbox}
\end{minipage}
\caption{The prompt that used to refine the explanations for grounding and segment visual objects}
\label{tab:refine}
\end{table*}

\section{Qualitative Steering Experiments}
\label{appendx:qualitative-steering}

\begin{table}[h]
\vspace{-1em}
\centering
\resizebox{0.5\linewidth}{!}{
\begin{tabular}{lllll}
\toprule
                          & Steering & Random &  &  \\
\midrule
GPT-4o Score & 6.36     & 2.02   &  &  \\
\bottomrule
\end{tabular}
}
\caption{GPT4 Score Evaluation of Steering Effects}
\vspace{-0.5em}
\label{tab:steering}
\end{table}

In~\cref{tab:steering}, we present the results of the steering evaluation for selected cases. Due to the high cost of large-scale steering evaluations, qualitative results are largely absent in the literature, including ~\citep{templeton2024scaling}. To address this, we take an initial exploratory step by using an LLM to assess steering examples, demonstrating a potential solution.

\section{More Steering Examples}

\begin{figure}
    \centering
    \includegraphics[width=1\linewidth]{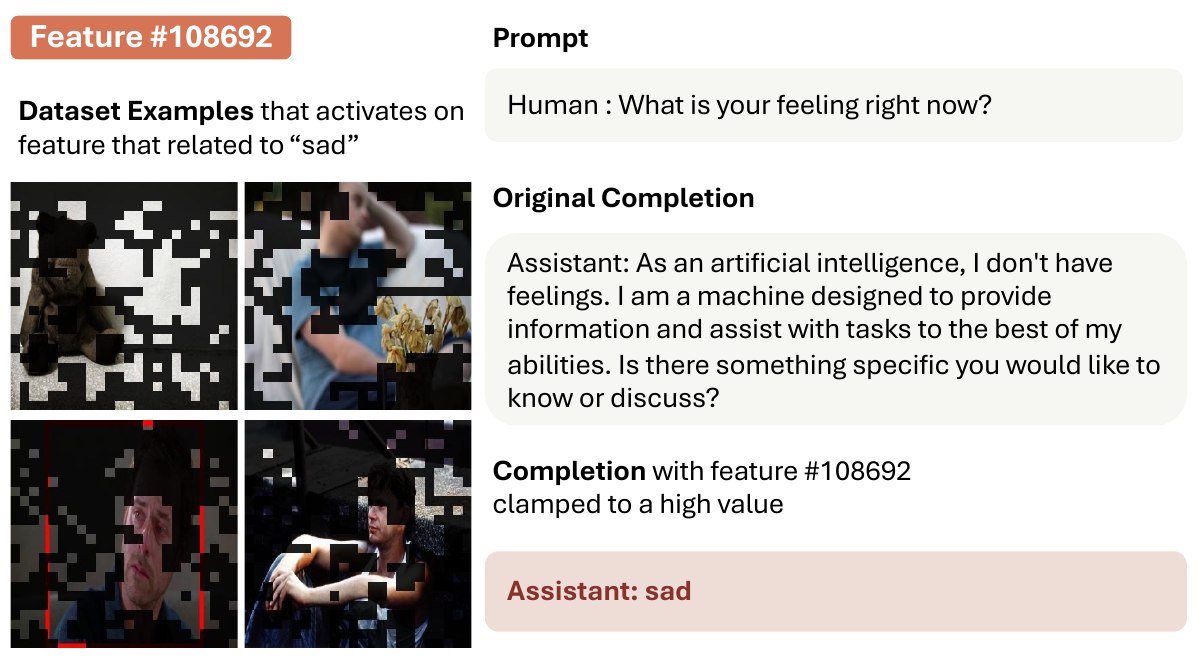}
    \caption{The feature that relates to sad. We probe and find out the feature that activated on ``sad''. By clamping this feature, we can enforce the model to share the feeling of sad}
    \label{fig:sad}
    \vspace{-4mm}
\end{figure}

\paragraph{Sad} We present a feature that may be related to the feeling of ``sadness'' and explore the potential for enabling the model to share emotional responses. After probing and confirming that the feature aligns with ``sadness,'' we investigate whether manipulating this feature could influence the model's reasoning to simulate emotional responses. To test this, we use a simple prompt, ``What is your feeling right now?'' and ask the assistant. Without steering, the model responds in a neutral, standard AI assistant tone, showing no emotion. However, when we clamp the ``sad'' feature to a high value, the model responds with ``sad'' as shown in~\cref{fig:sad}

In this section, we provide more steering examples that we discover during experiments. We perform a large scale steering on the 5000 size features subset we choose and then filtered some interesting examples here. In~\cref{fig:money}, the feature activates on money and when this feature is clamped to 50, the model output a story about saving funds and by a house. In~\cref{fig:speech}, when a feature relates to a feature that relate to speech, the model output a story about a man who is a speech master. In~\cref{fig:unix}, we found a feature that relate to unix/linux and its steering effect would output a story about Ubuntu. More interestingly, in~\cref{fig:chair}, though the model response on a visual ``chair'' object, when steering this feature, model would output a story relates to ``chairman'' instead of a ``chair''. Another example is that in~\cref{fig:keychain}, when steering this feature related to ``key'' or ``keychain'', the model output a story about developing some analytic software.

\section{CLIP-Score and IOU details}

We use Grounding DINO L~\citep{liu2023grounding} as our grounding module and SAM Huge~\citep{kirillov2023segany} as our segment module. The output from the interpretation pipeline is being refined into concise description by using the LLaMA-3.1-Instruct-8B~\citep{dubey2024llama3herdmodels}. We use ViT-B/32 CLIP model to generate embeddings and calculate the cosine similarity between the interpretations and the image. We calculate the IOU and the CLIP-Score using the top-5 activated images for each features. Due to the same limitation as illustrate in~\citep{templeton2024scaling,bricken2023monosemanticity}, we report the result on a 5000 subset of features with around 46684 images for caching the features' activations.

\begin{figure*}
    \centering
    \includegraphics[width=1\linewidth]{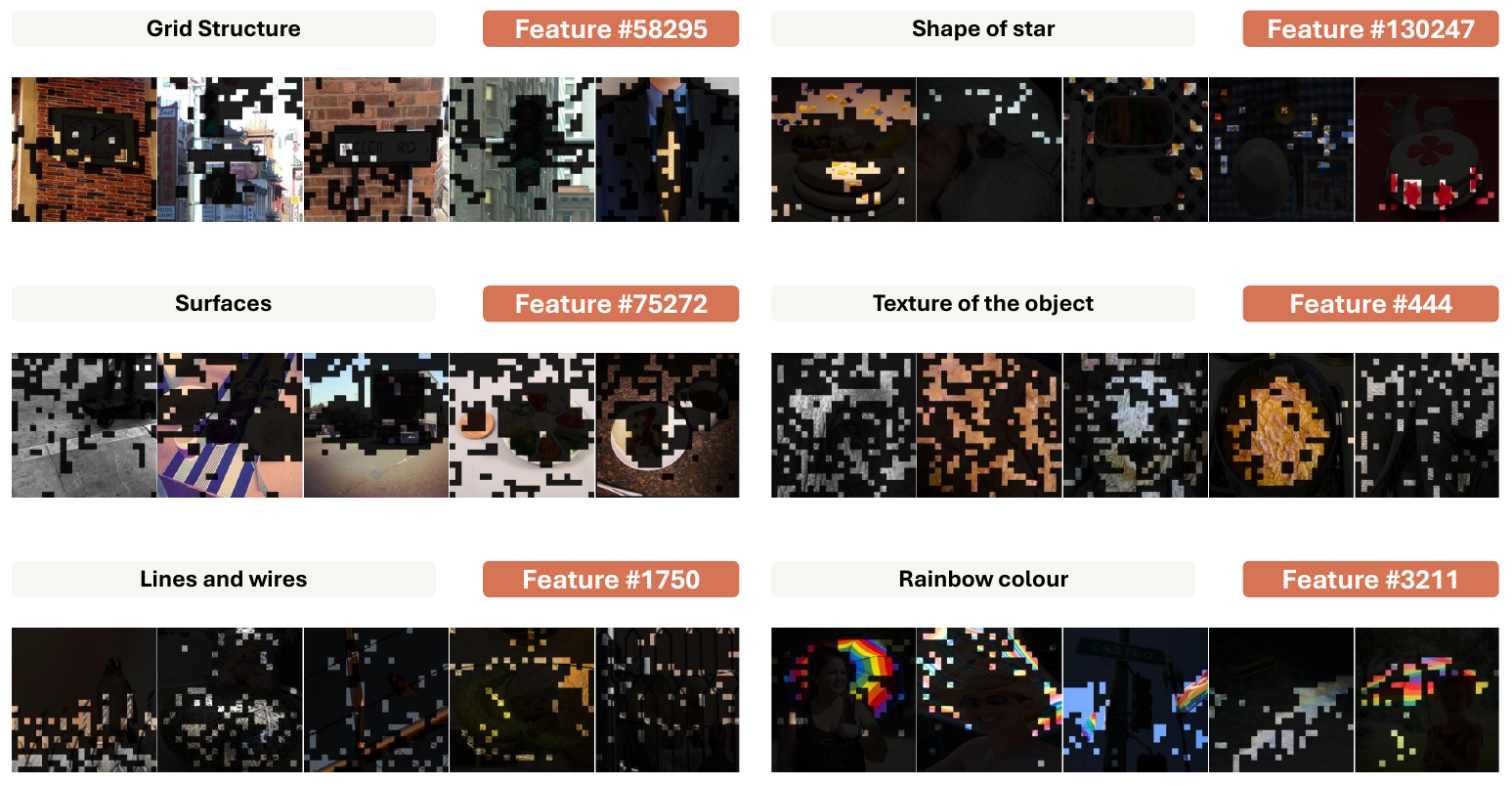}
    \caption{Low level features in the LMM. These features activate in most of the images and showcase the model's basic cognition and perception abilities.}
    \label{fig:low-level}
\end{figure*}

\begin{figure}
    \centering
    \includegraphics[width=1\linewidth]{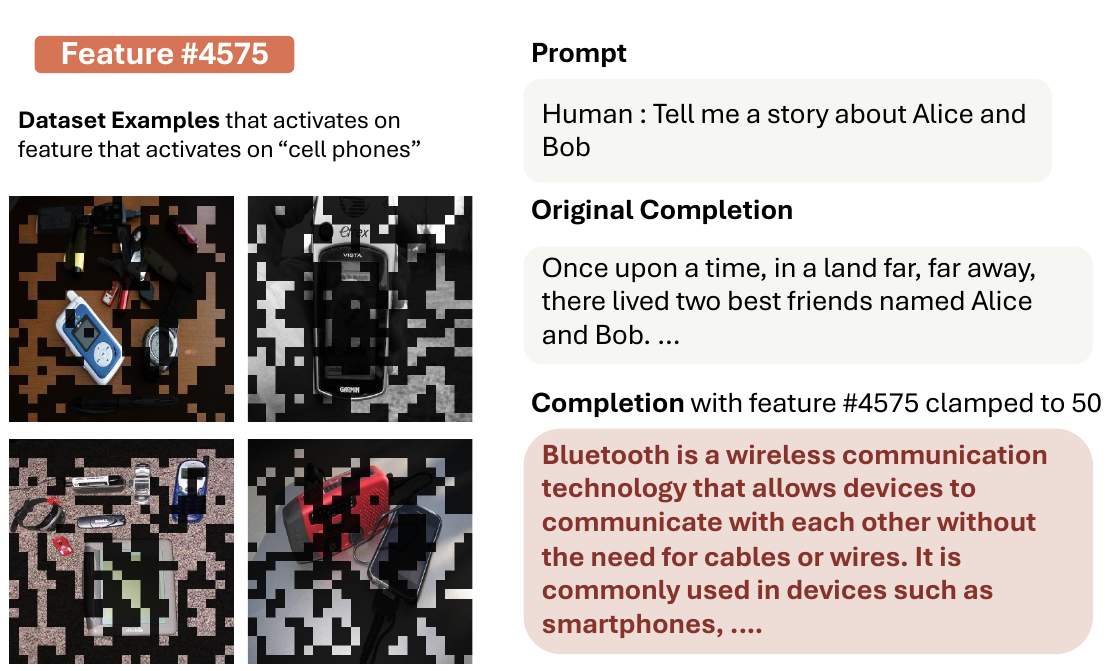}
    \caption{The feature related to money and its steering effect.}
    \label{fig:phone}
\end{figure}

\section{Feature Probing}

Due to the large number of features, identifying specific features of interest is challenging, and interpreting all available features before making a selection is impractical. Following~\citet{templeton2024scaling}, we also probe into the features of our SAE to identify several emotion-related features that may influence the model's perceived emotional responses. We first prepare an image representing a specific emotion, then select the top-k activated features for that image to run through our explanation and steering pipeline. From the output, we manually select the desired features and validate them through steering and activated examples. Unlike the approach in~\citep{templeton2024scaling}, which uses only the top 5 activated features, we found that a higher value of $k$ is preferable because a single image can contain many low-level visual features and diverse semantic information. In practice, we select $30 \leq k \leq 100$ and skip some of the top-activated values to exclude low-level visual features.

\section{Low Level Perception Features Examples}
\label{appendix:low-level}

We identify many low-level visual features from the model that differ from the text-based features in large language models (LLMs). These visual features are strongly activated across most images and represent the model's basic perceptual and cognitive abilities. In~\cref{fig:low-level}, we present examples of features activated by structure, shape, and color. In many of our probing trials, these features exhibit high activation levels and respond to various aspects of the images. We believe these features function as universal elements in how language-vision models (LMMs) understand the world.

\section{More Model comparison}

\begin{table}[h]
\centering
\resizebox{0.7\linewidth}{!}{
\begin{tabular}{lllll}
\toprule
            & IOU   & IOU(random) & CS    & CS(random) \\
\midrule
Qwen-2.5-VL & 26.67 & 0.06        & 27.99 & 18.22      \\
InstructBLIP-7B & - & -        & 28.01 & 17.71      \\
\bottomrule
\end{tabular}
}
\vspace{-1em}
\caption{Pipeline results on Qwen2.5-VL}
\label{tab:qwen-vl}
\vspace{-0.75em}
\end{table}

We provide a further experiment using Qwen-2.5-VL to prove the generalizability of our methods. As shown in ~\cref{tab:qwen-vl}

\section{Hallucination Steering Examples}

\begin{table}[h]
\centering
\vspace{-0.8em}
\resizebox{0.5\linewidth}{!}{
\begin{tabular}{llll}
\toprule
           & Better & Same & Worse \\
\midrule
HalluBench & 0.09   & 0.89 & 0.02  \\
\bottomrule
\end{tabular}
}
\vspace{-0.7em}
\caption{Hallucination Case study on 100 examples on Hallucination Bench with a single feature clamped at high value}
\label{tab:example}
\vspace{-1.1em}
\end{table}

We conduct a small-scale experiment on the Hallucination Bench by clamping irrelevant features, and the results are presented in Table~\ref{tab:example}. Among the 100 examples, clamping led to improved performance in 9 cases. Although this investigation is still in its early stages, we believe this approach shows potential for reducing hallucinations.

\end{document}